\documentclass[letterpaper]{article} 
\usepackage{main}  
\usepackage{times}  
\usepackage{helvet}  
\usepackage{courier}  
\usepackage[hyphens]{url}  
\usepackage{graphicx} 
\usepackage{natbib}  
\usepackage{caption} 
\usepackage{algorithm}
\usepackage{algorithmic}
\usepackage{amsfonts}
\usepackage{booktabs}

\usepackage{newfloat}
\usepackage{listings}
\DeclareCaptionStyle{ruled}{labelfont=normalfont,labelsep=colon,strut=off} 
\floatstyle{ruled}
\newfloat{listing}{tb}{lst}{}
\floatname{listing}{Listing}
\usepackage{amsmath}
\usepackage{adjustbox}
\usepackage{appendix}

\usepackage{color}
\definecolor{darkpink}{RGB}{231,84,128}

\title{

How Your Location Relates to Health: Variable Importance and Interpretable Machine Learning for Environmental and Sociodemographic Data
}

\author{
    Ishaan Maitra\textsuperscript{\rm 1}\equalcontrib, Raymond Lin\textsuperscript{\rm 1}\equalcontrib, Eric Chen\textsuperscript{\rm 1}, Jon Donnelly\textsuperscript{\rm 1}, Sanja \v{S}\'{c}epanovi\'{c}\textsuperscript{\rm 2}, Cynthia Rudin\textsuperscript{\rm 1}\
}
\affiliations{

    \textsuperscript{\rm 1}Duke University, North Carolina, United States\\
    \textsuperscript{\rm 2}Nokia Bell Labs, Cambridge, United Kingdom \\
    ishaan.maitra@duke.edu, raymond.lin@duke.edu,
    eric.y.chen@duke.edu,
    jon.donnelly@duke.edu,
    sanja.scepanovic@nokia-bell-labs.com,
    cynthia@cs.duke.edu
}

\begin{document}

\raggedbottom 

\maketitle

\begin{abstract}

Health outcomes depend on complex environmental and sociodemographic factors whose effects change over location and time. Only recently has fine-grained spatial and temporal data become available to study these effects, namely the MEDSAT dataset of English health, environmental, and sociodemographic information. Leveraging this new resource, we use a variety of variable importance techniques to robustly identify the most informative predictors across multiple health outcomes. We then develop an interpretable machine learning framework based on Generalized Additive Models (GAMs) and Multiscale Geographically Weighted Regression (MGWR) to analyze both local and global spatial dependencies of each variable on various health outcomes. Our findings identify NO2 as a global predictor for asthma, hypertension, and anxiety, alongside other outcome-specific predictors related to occupation, marriage, and vegetation. Regional analyses reveal local variations with air pollution and solar radiation, with notable shifts during COVID. This comprehensive approach provides actionable insights for addressing health disparities, and advocates for the integration of interpretable machine learning in public health.

\end{abstract}


\begin{links}
    \link{Code}{github.com/maitraishaan/VI-and-IML-MEDSAT}
    \link{Datasets}{mediatum.ub.tum.de/1714817}
\end{links}

\section{Introduction}

Recent increases in the prevalence of chronic diseases such as diabetes and hypertension \cite{acc:22} along with empirical evidence of the health risks associated with climate change \cite{obradovich2018empirical} highlight the pressing need to understand what environmental and sociodemographic features impact health. Only recently has a very detailed dataset \citep[MEDSAT,][]{MEDSAT} become available that integrates health determinants, including sociodemographic, environmental, and geographic features, with outcome-related prescription data as a proxy for health outcomes. It offers an unprecedented opportunity for exploring these relationships in close detail across England -- if we have the trustworthy tools to do it.

Prior work has utilized other integrated datasets to uncover relationships between environmental or sociodemographic variables and health outcomes, but these studies tend to 1) rely on a single variable importance method, 2) fail to measure spatially local relationships, and 3) do not create tools that provide extremely interpretable and auditable models. A group of recent works have focused on demonstrating how air pollution triggers respiratory and cardiovascular diseases \cite{am21, RazaviTermeh2021, Ravindra2023, Liu2024}. Other works have aimed to determine risk factors for depression and anxiety \cite{Qasrawi2022, Byeon2021, Elshawi2019}. These studies have a primary limitation: the lack of a \textit{comprehensive} dataset that would enable the disentangling of contributions from multiple factors to health outcomes. The lack of a large, complex dataset has been a challenge for environmental public health studies for years \cite{Comess2020}, but MEDSAT addresses this gap by including diverse environmental, sociodemographic, and health variables, enabling more nuanced analysis.

Newer works examining risk factors and mortality for COVID-19 \cite{Shao2021, Hu2024, Scarpone2020} have popularized the use of spatial modeling techniques to account for inherent spatial heterogeneity within geographic data. 
We draw inspiration from these works but highlight the need for comprehensive variable filtering to identify only the most relevant factors in large datasets, improving interpretability and actionable insights \cite{Elshawi2019}. To do this, our modeling uses a dual approach: global interpretability techniques that generalize over the entire population of England and local interpretability techniques that provide explanations at much smaller levels. This allows us to have very detailed insight into \emph{which} variables are important, and more crucially, \textit{where} they are important.

We develop a robust framework that integrates comprehensive importance ranking with interpretable machine learning techniques to effectively filter and analyze sociodemographic, geographic, and environmental variables. This approach is applied to key health outcomes such as diabetes, anxiety, hypertension, asthma, and depression, allowing us to examine their associations both locally and globally, across pre- and post-COVID-19 periods. For our variable filtering step, we first use knockoffs \cite{Barber_2015} to substantially shrink our variable set for downstream spatial models, and then extract the top 10 variables using an average of SHAP \cite{lundberg2017unified}, LOCO \cite{lei2018distribution}, Permutation Importance \cite{Altmann2010}, and Conditional Model Reliance \cite{fisher2019modelswrongusefullearning} rankings. Next, we use Generalized Additive Models (GAMs) \cite{hastie1990generalized} with a spatial smoothing tensor to identify the variables that are ``globally'' important across most geographic regions, and then use Multiscale Geographically Weighted Regression (MGWR) \cite{Oshan2019} and GAMs fit on each England Local Authority District (LAD) to identify local variations for the predictors.

Our study identifies key global variables associated with health outcomes, with NO2 consistently linked to asthma, hypertension, and anxiety across pre- and post-COVID-19 periods. We also uncover interesting outcome-specific predictors related to work, marriage, and vegetation. Through detailed spatial analyses using GAMs and MGWR, we highlight regional patterns and shifts in variables such as PM 2.5, revealing notable shifts during the COVID-19 period. This work serves as a blueprint for future public health research and policy-making, and also as proof of the importance of integrating interpretable machine learning models into public health. The potential impact is substantial in offering guided strategies to improve public health outcomes, identifying critical health determinants, and informing policy decisions. Collaboration with public health experts will be key to translating these insights into actionable strategies at both local and national levels.

\section{Related Work}

We use MEDSAT \cite{MEDSAT}, a comprehensive dataset created in 2023 to facilitate the exploration of the complex relationships between environmental factors, sociodemographic characteristics, and public health outcomes across England. It integrates various data sources:
\textbf{1) Medical Prescription Data}:
    Obtained from the National Health Service in England, these data include detailed information on outcome-related prescriptions covering various medical conditions such as diabetes, hypertension, asthma, and opioid use. We refer to the health outcome variable (such as diabetes or anxiety) as the total quantity of prescriptions related to that outcome.
\textbf{2) Sociodemographic Data}:
    Sociodemographic data were sourced from the England Office for National Statistics and includes information on population demographics, socioeconomic status, employment, education, and housing conditions.
\textbf{3) Environmental Data}:
    Environmental data include measurements of air quality (e.g., NO2, PM2.5), temperature, humidity, and other environmental factors.

Our work uniquely explores the MEDSAT data set, integrating existing spatial modeling techniques in our analysis of outcomes. \citet{Shao2021} used a spatial regression technique called geographically weighted regression (GWR) to predict COVID-19 cases in China based on population flow variables, geolocation, and socioeconomic data. GWR \cite{brunsdon1996geographically} models spatially varying relationships by allowing predictor coefficients to change across locations within a bandwidth, which is the count of neighboring points used for spatial smoothing. However, it applies the same spatial scale to all predictors, which can be limiting. Multiscale Geographically Weighted Regression (MGWR) \cite{Oshan2019} enhances GWR by estimating optimal bandwidths for each predictor. \citet{Hu2024} utilizes MGWR to assess the influence of mobility and health-related factors on COVID-19 mortality in the US.

While MGWR captures spatial heterogeneity at multiple scales, it struggles with non-linear relationships and correlated predictors. Generalized Additive Models (GAMs) \cite{hastie1990generalized} address these issues by extending linear models to allow for non-linear effects while maintaining interpretability through smooth, single-variable relationships. For instance, \citet{Scarpone2020} used GAMs to identify variables associated with COVID-19 at the county scale, and \citet{sun2023prediction} applied GAMs to predict respiratory disease mortality using air pollution and weather factors. We find benefits in leveraging both: MGWR is uniquely suited for revealing general spatial variations in predictor effects, identifying regions where a variable's relationship with the outcome differs significantly from that of neighboring areas, whereas GAMs are ideal for modeling the complex relationships within these regions. This combination enables a more thorough spatial analysis and addresses limitations of single-model approaches in prior studies.

A significant drawback to GAMs is that they can become too complex with many input variables, making them hard to interpret \cite{datadance2023gam}. The MEDSAT dataset contains 154 variables and 28,641 data points. Fitting a GAM on the full dataset is impractical due to the high number of variables, so we incorporate variable filtering via variable importance metrics to improve downstream model performance and interpretability. Notably, this step is non-trivial because many variable importance metrics exist, each with its own strengths and weaknesses.

We begin by leveraging SHAP \cite{lundberg2017unified}, a post-hoc explanation method for feature importance. SHAP values explain each feature's contribution by distributing the prediction value among the features based on their marginal contributions across all possible feature subsets. However, the correctness and existence of the baseline values used in SHAP are difficult to verify, which can lead to potential misestimation. SHAP can also be impacted by its approximation that is used to reduce computation, and by extrapolation issues that affect permutation methods \citep{HookerEtAl2021}. 


To address the potential issues with SHAP, we incorporate permutation importance \cite{Altmann2010}, conditional model reliance (CMR) \cite{fisher2019modelswrongusefullearning}, and LOCO \cite{lei2018distribution} as complementary methods. Permutation importance assesses the impact of feature shuffling on overall model performance, making it more robust to multicollinearity, but destroys the covariance between predictor variables, leading to out of distribution values after permutation. LOCO measures a variable's importance by its impact on model performance when left out, reducing importance inflation that can occur with other metrics. However, LOCO can consider an omitted variable unimportant when other variables are correlated with the omitted one, and requires fitting a new model for each variable. CMR evaluates a variable’s importance by measuring the model's reliance on its unique information without refitting -- specifically, the contribution of a feature that cannot be explained by the remaining correlated features. This ensures that the covariance structure between variables is preserved, preventing out-of-distribution issues. 

Given the scaling challenges associated with SHAP, LOCO, and CMR with 154 variables, we use knockoffs \cite{Barber_2015} as an initial filtering step. Knockoffs compare model performance with true features against synthetic knockoff versions, preserving variable associations without true outcome relationships. This method controls for false discoveries and reliably identifies a subset of important variables. However, knockoffs do not provide insight into relative importance rankings or capture non-linear/ interaction effects between the variables and the outcome. This is where the variable importance metrics are useful.

\section{Methodology}

\begin{figure}[t]
\centering
\includegraphics[width=0.65\columnwidth]{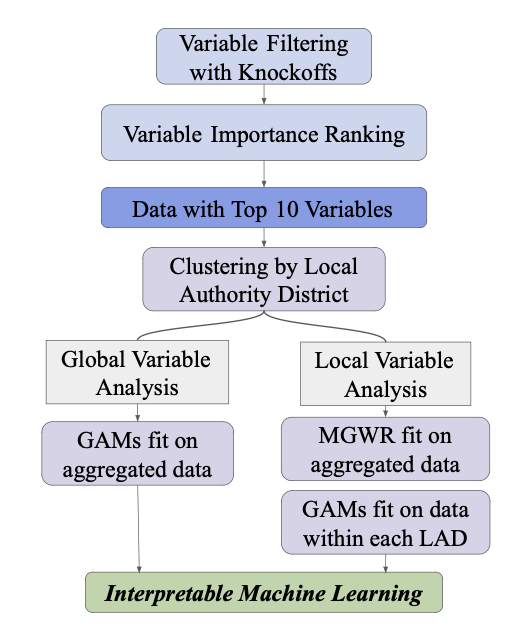} 
\caption{Overview of the full methodology pipeline.}
\label{figmethod}
\end{figure}

\subsection{Step 1: Variable Filtering with Knockoffs}

We reduce the high-dimensional MEDSAT data to the most relevant predictors to avoid overfitting and dilution of predictive power in downstream models. As discussed, we first employ the knockoffs filter to control the false discovery rate (FDR), thereby improving the reliability of our methods. Knockoffs create control variables \(\tilde{\mathbf{X}}\) replicating the correlation structure of the original variables \(\mathbf{X}\). We compute the statistic \(W_j = |Z_j| - |\tilde{Z}_j|\), where \(Z_j\) and \(\tilde{Z}_j\) are importance measures of \(\mathbf{X}\) and \(\tilde{\mathbf{X}}\), respectively. Variables with \(W_j\) exceeding a threshold \(T\) are kept, where \(T\) is chosen to control the FDR at a desired level \(q\). We tune parameters such as the FDR and knockoff ``feature statistic'' using 5-fold cross-validation.

Next, we determine the optimal hyperparameters for an XGBoost \cite{chen2016xgboost} regressor on the post-knockoffs variables. We use randomized grid search cross-validation to optimize the \(R^2\) score, and employ 5-fold stratified CV (over parameters including number of estimators, max depth, and learning rate) to maintain a distribution similar to the England Local Authority Districts (more on this later), reserving 10\% of the data for test evaluation. We chose XGBoost due to its reliability and demonstrated proficiency in capturing complex spatial effects \cite{li2022extracting}.

Using these hyperparameters, we fit a set of 50 XGBoost models, each initialized with a different random seed, to create a diverse set of models. By fitting multiple XGBoost models, we aim to draw several samples from a Rashomon set \cite{fisher2019modelswrongusefullearning,SemenovaRuPa2022} -- a collection of equally accurate models that differ slightly in their learned relationships with the data. To enhance this, we applied rejection sampling, discarding models with poor performance relative to the best models. This diversity helps to ensure that our variable importance analysis reflects robust, consistent patterns rather than the idiosyncrasies of a single model.

\subsection{Step 2: Variable Importance Ranking}

As the second step in our variable filtering pipeline, we evaluate the variable importance of each XGBoost model using several metrics. \textbf{(1) Permutation Importance}: 
For each feature \( j \), we measure the change in model performance when the feature values are randomly shuffled. Let \( s \) be the baseline score (e.g., \( R^2 \) or accuracy) of the model. The feature importance \( I_j \) is calculated as the difference between the baseline score and the score after permutation $ s_{\text{perm}(j)} $: $ I_j = s - s_{\text{perm}(j)} $.
\textbf{(2) SHAP Values}:
SHAP values provide feature importance based on cooperative game theory. For each feature \( j \), the SHAP value \( \phi_j \) represents the average contribution of the feature across all possible subsets of features. 
\textbf{(3) LOCO Importance}:
LOCO importance involves removing each feature \( j \) one at a time, refitting, and measuring the change in model performance. Given a baseline score \( s \), the feature matrix \( X_{-j} \) is created by excluding feature \( j \), and the model is refit to obtain a new score \( s_{-j} \). The importance \( I_j \) is calculated as: $I_j = s - s_{-j}$.
\textbf{(4) Conditional Model Reliance}:
CMR quantifies the unique importance of a feature \( X_j \) by evaluating the change in model performance when the feature is imputed using the other observed covariates. Using a baseline model trained on the original dataset, predictions are compared against predictions by the baseline using a dataset where \( X_j \) is imputed. 

Using the four variable importance metrics calculated for each of the post-knockoffs variables, we compute a mean rank. Using the mean rank allows outliers, such as variables with high LOCO importance but low SHAP importance, to influence feature selection, thus preserving useful signals that might otherwise be overlooked. We use the top 10 most important variables per outcome -- these are the variables used in downstream models -- to strike a balance between capturing the most influential predictors and maintaining model simplicity. We found that adding more variables to our models added little to no additional predictive performance.

\subsection{Step 3: Geographic Clustering by Local Authority District}

The 28,641 data points in MEDSAT are grouped into Local Authority Districts (LADs), which define a geographical boundary for British census units (UK Office of National Statistics 2021). Each LAD encompasses multiple Lower Layer Super Output Areas (LSOAs) and represents a region with relatively homogeneous socioeconomic and demographic characteristics. As a result, we use LADs as our modeling unit because they capture spatial dependencies and variations in our outcome variables over regions that present an appropriate administrative partitioning for enacting policy. This also reduces our computational complexity, which is an issue for GAMs and MGWR on large datasets. 

\subsection{Step 4: Global Variable Analysis with GAMs}

Due to the non-linear relationships that are common in environmental and health data \cite{Zhang2022}, GAMs are useful for analyzing both the magnitude and shape of relationships between filtered variables and outcome variables. 

Before fitting the model, we first make sure that none of the 10 ranked variables per outcome are highly correlated with each other using a threshold of 0.8. For instance, there are several different variants of the ``percent divorced'' variable (percent divorced, percent divorced or partnership dissolved, etc.) in the dataset and we do not wish to keep all of them. For any correlated pairs, we remove the less important variable using our prior ranking. Second, we aggregate the data by each LAD, using the centroid point of each LAD's data as the aggregated coordinate and take the mean of all predictor and outcome variables. Then, we fit a GAM across the centroids with the following form -- including a tensor product between the x and y-coordinates to account for spatial heterogeneity:
\[
    y = \beta_0 + s_1(X_1) + s_2(X_2) + \ldots + s_n(X_n) + tensor(c_{x_1}, c_{x_2})
    \]
    where \(y\) is the outcome variable (e.g., diabetes quantity per capita), $\beta_0$ is the bias term, \(s_i(X_i)\) is the smooth function applied to the \(i\)-th predictor, and $tensor(c_{x_1}, c_{x_2})$ represents the smooth spatial effect term between the $x_1$ and $x_2$ coordinates for the LAD centroid.

For each model, we perform a grid search over the lambda and n\_splines smoothing parameters, maximizing the generalized cross validation score (GCV). Lambda controls the level of L2-regularization, while the number of splines controls the number of basis functions for each smooth function. 

To explore the relationship between each predictor variable and the outcome, we visualize the shape functions. By examining the direction and strength of the relationships depicted in the shape functions, along with 95\% confidence intervals surrounding the estimates, we can reason if a variable is a strong global predictor of a health outcome.

This global GAM model is used only to gain a general understanding of which variables might be important overall. We switch to local models next.

\subsection{Step 5: Modeling Local Variations with MGWR}

We utilize a two-pronged approach combining MGWR and GAMs. We used MGWR to provide regions of interest where the model coefficients are relatively constant. Some small regions have outlier-like coefficients, and are considered as their own regions of interest. After this, we provide a GAM for each region to predict outcomes for the locations within that region. (These local GAMs are different than the global GAM described in the last step.)

MGWR is a local linear regression model that, for each predictor, fits unique coefficients at LAD centroids using a kernel-based weighting scheme that accounts for spatial variations and dependencies. MGWR improves on traditional GWR models by allowing the bandwidth $b_k$, which determines the local neighborhood for weighting each regression point, to vary across different predictors. In other words, $b_k$ controls the degree of smoothing -- a smaller $b_k$ means the smoothing neighborhood is smaller.

The local regression equation for each LAD centroid \( i \) is formulated as follows:
\[
y_i = \beta_{0i} + \sum_{k=1}^{p} \beta_k(u_i, v_i, b_k) x_{ik} + \epsilon_i
\]
where \( y_i \in \mathbb{R} \) represents the health outcome value for location \( i \), \( (u_i, v_i) \in \mathbb{R}^2 \) are the coordinates of location \( i \), \( \beta_{0i} \in \mathbb{R} \) is the spatial bias term, \( \beta_k(u_i, v_i, b_k) \in \mathbb{R} \) is the spatially varying coefficient for the $k$-th predictor variable smoothed across bandwidth $b_k$, \( x_{ik} \in \mathbb{R} \) denotes the value of the \( k \)-th predictor variable at location \( i \), \( \epsilon_i \in \mathbb{R} \) is the Gaussian error term for location \( i \), and \( b_k \in \mathbb{R} \) is the optimal bandwidth for predictor \( k \), determined via cross-validation.


\begin{table*}[!h]
        \centering
        \LARGE
        \resizebox{\textwidth}{!}{
        \begin{tabular}{lllllllllll}
             \toprule
             \textbf{Feature} & \textbf{Permutation} & \textbf{SHAP} & \textbf{LOCO} & \textbf{CMR} &
             \textbf{Feature} &  \textbf{Permutation} & \textbf{SHAP} & \textbf{LOCO} & \textbf{CMR} \\
             \midrule
             NO2 & 1 & 1 & 1 & 1 &
             NO2 & 2 & 5 & 1 & 1 \\
             
             Thermal radiation & 3 & 2 & 7 & 5 &
             Thermal radiation & 1 & 2 & 7 & 4 \\
             
             Skin reservoir content & 2 & 3 & 3 & 3 &
             PM2.5 & 4 & 1 & 8 & 2 \\
             
             PM2.5 & 7 & 5 & 2 & 2 &
             Snow depth & 3 & 4 & 3 & 3 \\
             
             Snow depth & 4 & 4 & 6 & 4 &
             Skilled trades occupation & 6 & 3 & 2 & 5 \\
             
             Pct commute metro rail & 8 & 8 & 9 & 6 &  
             Total solar energy absorption & 9 & 8 & 12 & 10 \\
             
             Surface latent heat flux & 6 & 6 & 13 & 7 & 
             Total precipitation & 8 & 14 & 9 & 13 \\
             
             Pct divorced & 9 & 9 & 5 & 12 & 
             Skin reservoir content & 5 & 7 & 17 & 9 \\
             
             Total solar energy absorption & 5 & 7 & 10 & 10 &
             Total evaporation sum & 7 & 9 & 14 & 8 \\
             
             Total evaporation sum & 10 & 14 & 11 & 8 &  
             Travel for work $\geq$ 60 km & 15 & 16 & 6 & 6 \\
             
        \bottomrule
        \end{tabular}}
        \caption{Variable importance ranks for anxiety (left) and diabetes (right) for the top 10 variables.}
        \label{tab:table1}
\end{table*}

The vector of locally estimated coefficients \( {\beta}_i \) is derived with the weighted least squares problem:

\[
\hat{{\beta}}_i = ({X}_i^T {W}_i {X}_i)^{-1} {X}_i^T {W}_i {y}_i
\]
where \({W}_i \) is the diagonal matrix of weights \( w_{il} \). The weights \( w_{il} \) are calculated based on the Euclidean distance between locations \( i \) and \( l \) if \( d_{il} \) falls within bandwidth \( b_k \).

\subsection{Step 6: Local Variation Analysis with Local GAMs}

We use the MGWR to manually identify regions of interest where coefficients are relatively constant. We will then create a separate GAM for each LAD to explicitly model relationships at the local level. We will then compare these local GAMs between the regions. 

In more detail, after the regions of interest are chosen, we fit a distinct GAM to the data within each LAD. To enhance our spatial analysis, we incorporate a form of spatial smoothing that emulates the weighted local regression bandwidth used in MGWR. This is achieved by introducing a predefined weight matrix, derived from an exponential decay function based on the distances between neighboring LADs, which adjusts the model fit for each LAD. These predefined weights modify the objective function, where each LAD's residual is multiplied by its corresponding weight, ensuring that closer neighboring LADs have a greater impact. 
We incorporate grid search over the number of splines and regularization terms, maximizing GCV.

This approach allows us to summarize the local GAMs for each LAD using the regions created from MGWR.

\section{Results}

\subsection{1. Environmental variables rank higher than sociodemographic variables.}

Our variable importance ranks the top 10 variables for each health outcome. In the case of anxiety, the most important variables are NO2, solar radiation, skin reservoir content, particulate matter, and snow depth (see Table \ref{tab:table1}). These environmental variables consistently rank high across multiple health outcomes, highlighting their substantial association. 

To compare with the MEDSAT paper, we examine their SHAP values from a LightGBM model, where NO2, England residency (10+ years), thermal radiation, work from home, and very good health were the top variables for anxiety \cite{MEDSAT}. In our analysis, NO2 remains a top predictor, while the others are knocked off. Climate factors such as solar and thermal radiation positively correlate with mental health \cite{Wang2023}. Skin reservoir content, or water content in the vegetation canopy \cite{era5-land}, reflects broader environmental conditions affecting air quality and climate. PM2.5 is associated with neurotoxicity and mental health issues \cite{Potter2021}.

\begin{table}[h!]
    \centering
    \resizebox{\columnwidth}{!}{\begin{tabular}{llllll}
        \toprule
         \textbf{Feature} & \textbf{Permutation} & \textbf{SHAP} & \textbf{LOCO} & \textbf{CMR} \\
         \midrule
         NO2 & 1 & 2 & 1 & 1 \\
         Skin reservoir content & 5 & 4 & 3 & 3 \\
         Thermal radiation & 2 & 3 & 5 & 6 \\
         Snow depth & 4 & 6 & 4 & 5 \\
         PM2.5 & 7 & 5 & 6 & 2 \\
         Net annual income & 9 & 8 & 2 & 4 \\
         White ethnicity & 3 & 1 & 21 & 14 \\
         Pct commute by metro rail & 10 & 11 & 8 & 11 \\
         Total precipitation & 8 & 14 & 9 & 13 \\   
         \bottomrule
    \end{tabular}}
    \caption{Anxiety variable importance and ranks for the top 10 variables after reordering by mean rank.}
    \label{tab:table2}
\end{table}

While existing literature often emphasizes sociodemographic variables \cite{MEDSAT, Gergov2023, Qiu2023, Cheah2020}, our analysis suggests that environmental factors offer unique predictive power in explaining outcome-related prescriptions. To further investigate, we reinserted key demographic variables such as population density, age, and ethnicity into our anxiety model. The adjusted model showed similar top variables, indicating that environmental variables remain strong predictors (see Table \ref{tab:table2}).

However, this interpretation requires caution. Although environmental factors provide unique information not captured by sociodemographic variables, it is also clear that individual sociodemographic variables, such as ethnicity, still contribute valuable information. For instance, ``white ethnicity'' shows high importance in SHAP due to its interactions with other variables but is low in importance under LOCO and CMR analysis, likely because it can be inferred from other factors. Therefore, while environmental variables do offer distinct predictive insights, the lower rank of sociodemographic variables could be partly due to their high inter-correlation and indirect relationship with health outcomes through environmental exposures. For example, while income and ethnicity may not offer unique information about anxiety, they can impact living conditions, such as air pollution exposure, which is a strong predictor of anxiety.


\subsection{2. NO2 consistently emerges as a global predictor for asthma, hypertension, and anxiety.}

\begin{figure}[h!]
\centering
\includegraphics[width=0.85\columnwidth]{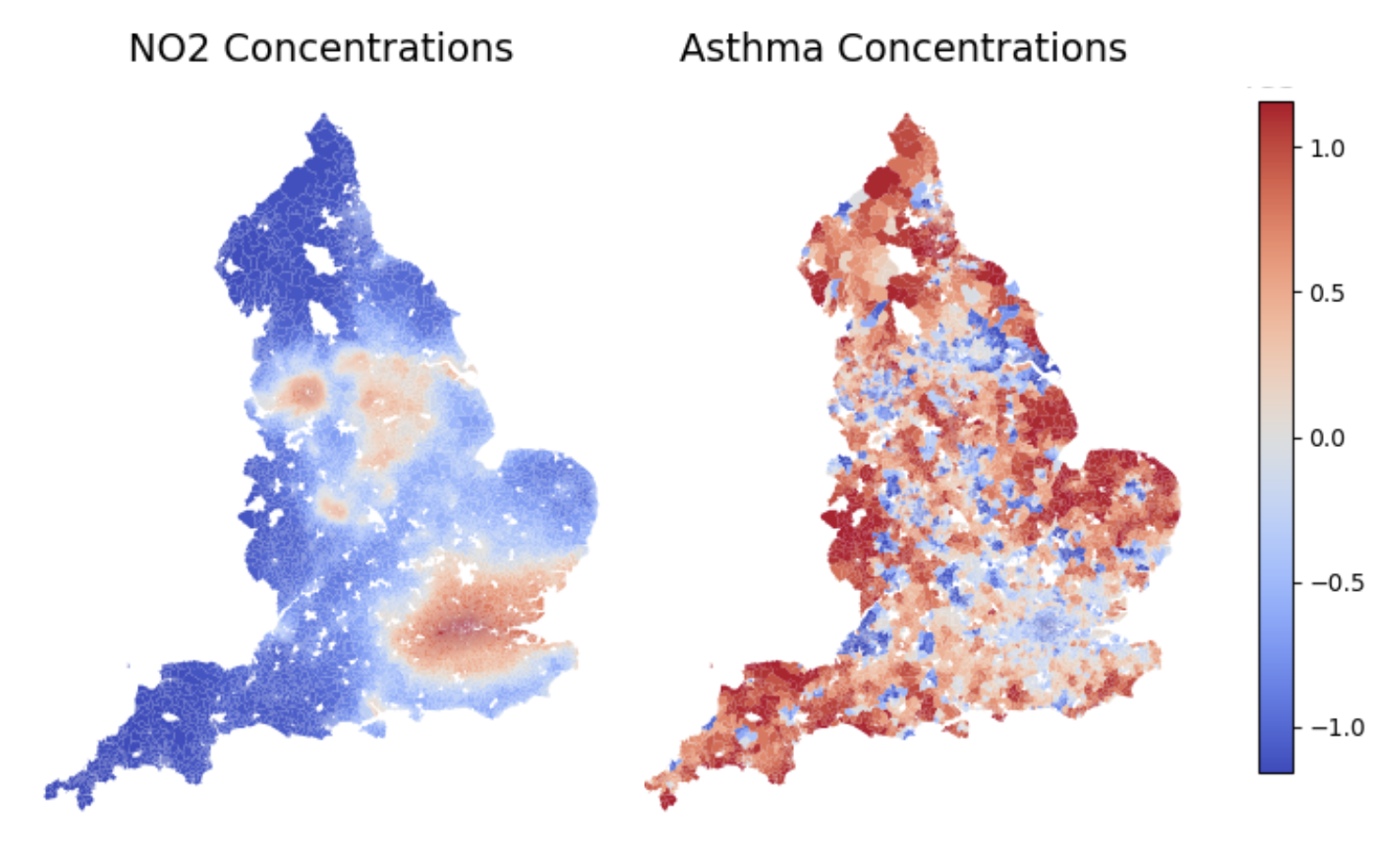} 
\caption{On left, NO2 concentrations. On right, asthma concentrations, 2019. Red regions have higher values.}
\label{fig_concentrations}
\end{figure}

Using our global GAMs, \textbf{NO2 consistently emerged as a global predictor} for prescription prevalence of asthma, hypertension, and anxiety, with narrow confidence intervals that do not overlap zero. Interestingly, despite the well-documented links between air pollution and respiratory issues \cite{am21}, our analysis reveals a \textbf{negative correlation} between NO2 levels and these health outcomes. This unexpected finding -- which is also found in the MEDSAT paper \cite{MEDSAT} -- is particularly evident in areas around London, which have the highest NO2 concentrations in England (see Figure \ref{fig_concentrations}) but show a weak negative relationship with asthma prescriptions based on our MGWR model in Figure \ref{fig_skilled}.

\begin{figure}[t]
\centering
\includegraphics[width=0.73\columnwidth]{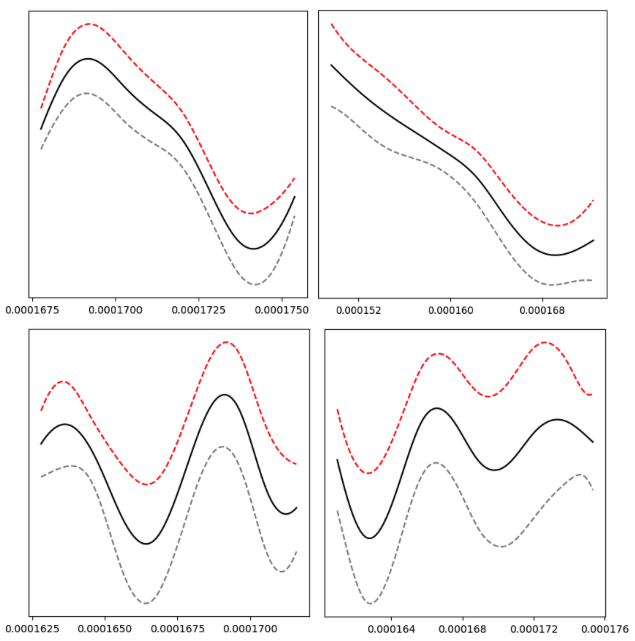} 
\caption{Shape Functions with 95\% CIs for NO2 with Asthma. On top, Westminster (left) and Lambeth (right). On bottom, Tower Hamlets (left) and Camden (right).}
\label{fig_NO2PDPS}
\end{figure}

To investigate this anomaly, we examined Local GAMs for London's 68 LADs, but with one adjustment: removing spatial smoothing to account for significant NO2 variation within the city (ranging from $1.14\times 10^{-4}$ to $1.71\times 10^{-4}$; across England, NO2 ranges from $6.0 \times  10^{-5}$ to $1.71 \times  10^{-4}$) and ensuring each LAD was considered independently. Our analysis reveals that the few districts with pronounced negative correlations were within the 2019 Ultra Low Emission Zones (ULEZ) boundary, an area encompassing London that charges non-compliant vehicles and thus averted a large amount of toxic air pollution in areas with pedestrians. In the first ten months, the implementation of ULEZ reduced roadside NO2 concentrations by 37\% \cite{London2020ULEZ}. As shown in Figure \ref{fig_NO2PDPS}, Westminster and Lambeth -- both in the ULEZ and two of the highest NO2 areas, with Westminster the highest -- show distinctly negative correlations and narrow 95\% CIs, while very nearby regions with high NO2 just outside the ULEZ, such as Hamlets and Camden, exhibit a much weaker relationship.

By integrating both global and local analyses, our pipeline uncovers critical insights, such as the nuanced effects of vehicular NO2 emissions, that would be missed with a purely global approach. Local GAMs help explain the confusing negative correlation between NO2 and asthma observed in our previous models, particularly in London, the region with the highest NO2 levels. The few areas with distinctly negative correlations were almost entirely within the ULEZ, where the relationship is explainable by the \textit{dramatic reductions in vehicular NO2 emissions, despite high overall NO2 levels}. In contrast, neighboring regions showed no such negative relationship, with some having slightly positive correlations. Spatial smoothing in MGWR and Global GAMs is the culprit: it spreads the strong negative correlation in the ULEZ to adjacent, but significantly different areas. Overall, our findings underscore the need to reduce vehicular emissions -- a direct trigger of asthma \cite{Kelly2021} -- through local interventions, suggesting that expanding ULEZ-like zones could be an effective strategy.

\subsection{3. Occupation, length of residence, vegetation, and marriage are outcome-specific global predictors.}

\begin{figure}[t]
\centering
\includegraphics[width=0.8\columnwidth]{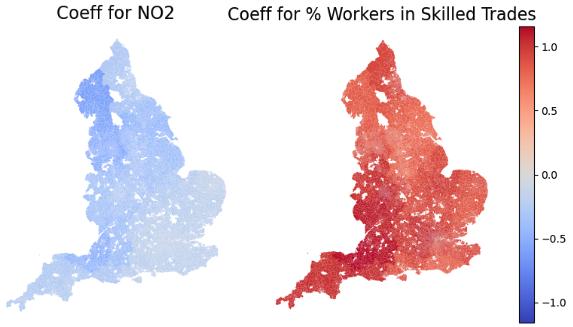} 
\caption{MGWR Coefficients for NO2 with Asthma (left), \% Workers in Skilled Trades with Diabetes (right), 2019.}
\label{fig_skilled}
\end{figure}

We determine four categories of other globally important variables that are more specific to an individual outcome: occupation, length of residence, vegetation, and marriage.
\textbf{Percent workers in skilled trades was a strong predictor of diabetes} prescriptions (see Figure \ref{fig_skilled}), but was not ranked post-filtering with other outcomes. \citet{Shockey2021} uncovered several occupation-related risk factors for diabetes: obesity, shift work, and job-related stress. Skilled trade workers -- for instance, welders, carpenters, and electricians -- exhibit several of these at risk factors, and can be multiple times more likely to develop Type 2 diabetes than occupations such as computer scientists or teachers \cite{Carlsson2020}. The proportion of residents as full-time workers also had a strong positive correlation with anxiety-related prescriptions across most of England except London. 

The proportion of residents with 10+ years in England and skin reservoir content show strong negative relationships with depression and hypertension prescriptions, respectively. The benefits of long-term residency can be attributed to improved social integration, better healthcare access, and more stable living conditions. On the other hand, the negative relationship of skin reservoir content suggests that healthier environmental conditions may help prevent chronic health conditions. 

\textbf{Percent married, percent divorced, and percent widowed are strong} global predictors for hypertension, depression, and anxiety prescriptions, respectively. This is well corroborated by prior research: \citet{Varghese2023} finds that married couples are likely to share hypertension, \citet{Menaghan1986} notes that newly divorced individuals are more depressed than their married counterparts, and \citet{Kristiansen2019} highlights the increased prevalence of depression and anxiety for widowed people. 

\subsection{4. PM2.5 has large local variations and a substantial
shift with COVID.}

\begin{figure}[t]
\centering
\includegraphics[width=0.67\columnwidth]{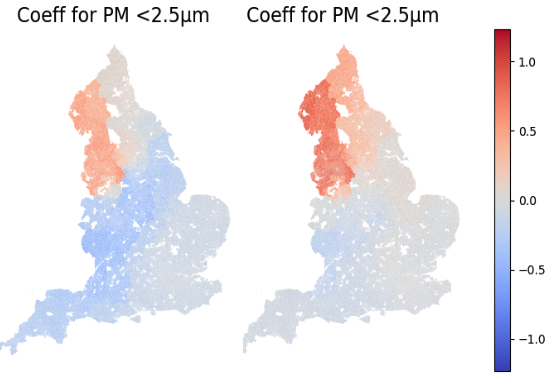} 
\caption{MGWR Coefficients for PM2.5 with Diabetes in 2019 (left) and 2020 (right).}
\label{fig_pm2.5.1}
\end{figure}

Particulate matter (PM 2.5) was extremely interesting as it showed substantial variations between LADs, and also had notable changes during COVID. For diabetes, we see in Figure \ref{fig_pm2.5.1} that areas in northwest England (Lancaster, Liverpool, Cheshire, Manchester) had distinctly positive correlations between PM2.5 and diabetes-related prescriptions, while the relationships in other regions were mostly near-zero or weakly-negative. However, after 2020, the northwest became more positively correlated while the rest of the country became almost entirely near-zero, despite a $\approx$15\% decrease in actual PM2.5 levels in the high-correlation LADs
This counter-intuitive result can potentially be explained by other factors -- particularly, the development of less healthy lifestyles during COVID and a 6.3\% increase in anxiety-related prescriptions in the $\geq 0.5$ coefficient LADs. \citet{Chien2016} finds that patients with anxiety have a higher incidence of diabetes.

\begin{figure}[h!]
\centering
\includegraphics[width=0.7\columnwidth]{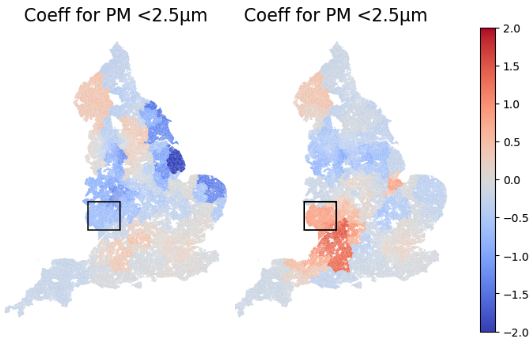} 
\caption{MGWR Coefficients for PM2.5 with Hypertension in 2019 (left) and 2020 (right).}
\label{fig_pm2.5.2}
\end{figure}

We see a dramatic regional shift with hypertension in Figure \ref{fig_pm2.5.2}, especially in the boxed region, which consists of Solihull, Worcester, and Wychavon. During COVID, all three regions have MGWR coefficients shift from $\le -0.5$ to $\ge 0.5$. However, analysis of the local GAMs shows that the individual splines for Worcester and Wychavon are noisy and likely impacted by the spatial smoothing in MGWR, while Solihull's shape function shows a stronger positive correlation. This contradiction between MGWR and Local GAMs is why we use both -- while MGWR highlights interesting regional variations, Local GAMs pinpoint specific LADs that should be targeted for policy intervention.

\subsection{5. Solar radiation is a crucial predictor of anxiety prescriptions -- but very dependent on region.}

\begin{figure}[t]
\centering
\includegraphics[width=0.45\columnwidth]{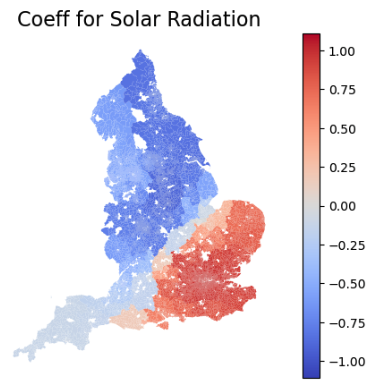} 
\caption{MGWR Coefficients for Net Solar Radiation with Anxiety, 2019.}
\label{fig_solar}
\end{figure}

Figure \ref{fig_solar} illustrates that the coefficient of net surface solar radiation with anxiety varies significantly between southeast (around London) and northern England. The strong coefficients across the map indicate a substantial correlation, but with opposite effects in these regions.

The southeast, receiving more solar radiation on average, may experience discomfort with further increases, unlike the colder northern regions that might benefit from it. Additionally, in densely populated areas of the southeast, higher solar radiation exacerbates the Urban Heat Island effect, increasing heat-related stress and anxiety, particularly in regions with minimal green space \cite{Keat2021}. Comparing the east and west coasts, we also note that waters off the east coast of England tend to be warmer than those off the west, meaning temperatures on the east coast tend to be higher even at the same latitude. With global warming likely to raise solar radiation levels further, our analysis highlights \textbf{southeast England as being at risk for rising anxiety} in the coming years.

\section{Conclusion}

Our analysis provides insights into the spatial and temporal dynamics of health outcome-related prescription prevalence across England. A  strength of our methodology lies in its rigorous variable importance process, ensuring that our models focus on the most relevant predictors. By leveraging MGWR and local/global GAMs, we effectively identify globally important variables while capturing regional variations both geographically and across pre- and post-COVID periods. This approach reveals associations between factors such as air pollution with health outcomes, offering valuable insights for both local and global policy decisions. The validity of our approach is underscored by its ability to highlight well established relationships, such as divorce correlating with depression. 
This framework also serves as a practical guide for future research and policy-making. 
Collaboration with public health experts will be essential to effectively implement these insights into strategies that improve public health.

\clearpage

\appendix

\bibliography{LaTeX/main}

\clearpage

\end{document}